\documentclass[11pt]{article}

\usepackage[preprint]{acl}

\usepackage{times}
\usepackage{latexsym}
\usepackage[T1]{fontenc}
\usepackage[utf8]{inputenc}
\usepackage{microtype}
\usepackage{inconsolata}
\usepackage{graphicx}
\usepackage{booktabs}
\usepackage{amsmath,amssymb}
\usepackage{xcolor}

\title{Do VLAs Really Follow Instructions? Diagnosing How-Level Instruction Following in Vision-Language-Action Models}
\newcommand{\proposed}{\textsc{SIMPLER-IN}}
\author{Sangwu Park \quad Wonjoong Kim \quad Yeonjun In \quad Chanyoung Park\thanks{Corresponding author.} \\
  KAIST \\
  \texttt{\{sangwu.park, wjkim, yeonjun.in, cy.park\}@kaist.ac.kr}}

\begin{document}
\maketitle

\begin{abstract}
Vision-language-action (VLA) models are expected to follow instructions like their language backbones, yet prior work has focused on What-level instructions that specify the task rather than How-level instructions that specify how to carry it out. Through our diagnoses, we find that How-level instructions do not help current VLAs and lower task success. We hypothesize that pairing each training episode with a single instruction limits How-level instruction following, and show that training with segment-level enriched instructions makes actions respond to such instructions without consistently raising task success. To measure How-level instruction following beyond standard benchmarks, we introduce \proposed, where success requires conditions given only in the instruction. It shows that the gained responsiveness does not extend to demanding or out-of-distribution How-level instructions. How-level instruction following thus remains an open challenge.\footnote{Our code and data are available at \url{https://github.com/sangwu99/how-level-vla-Arxiv}.}

\end{abstract}

\section{Introduction}
\label{sec:introduction}

Completing a task specified by a language instruction does not necessarily mean following instructions on how to perform it. For example, ``put the carrot on the plate'' specifies \emph{what} a robot should accomplish, whereas ``grasp the carrot by its green end'' specifies \emph{how} to accomplish it. This distinction is particularly important for vision-language-action (VLA) models, which generate robot actions from visual observations and language instructions~\citep{brohan2023rt, kim2024openvla, black2024pi_0, intelligence2025pi_, bjorck2025gr00t}. Following instructions about \emph{how} to act would enable VLAs to incorporate corrections, refine actions, and adjust their behavior through language, as widely practiced in large language models (LLMs) and vision-language models (VLMs)~\citep{ouyang2022training, madaan2023self, shinn2023reflexion, lee2024volcano, cheng2025vision, driess2023palm}. 

\begin{figure}[!t]
    \centering
    \includegraphics[width=\linewidth]{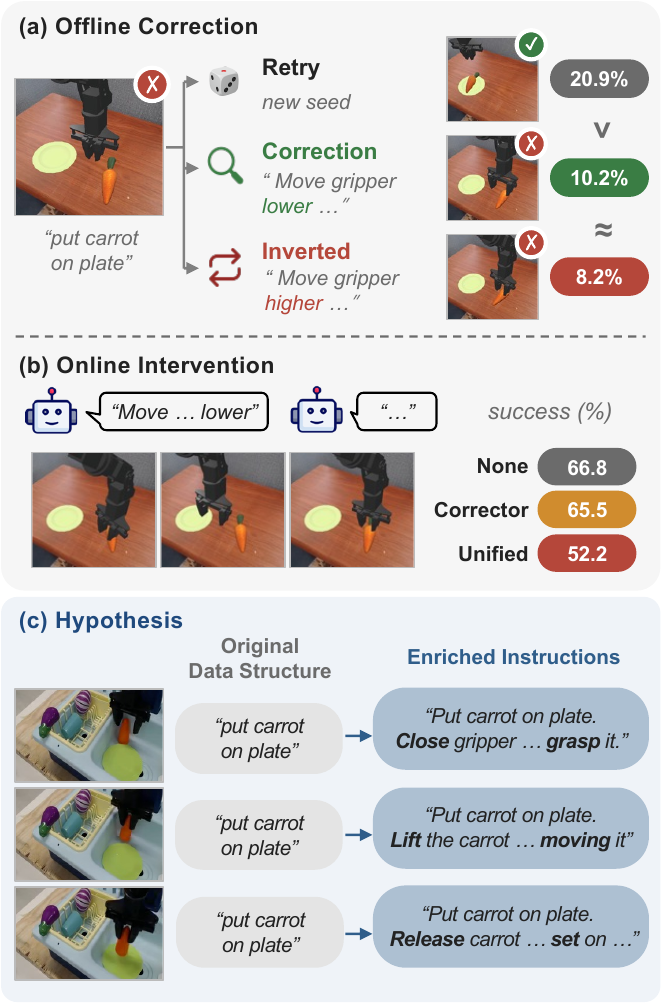}
    \caption{Overview of this work. (a), (b): Our diagnoses show that VLAs do not clearly follow How-level instructions. (c): Examples of enriched instructions.}
    \label{fig:teaser}
\end{figure}

We distinguish these capabilities as \emph{What-level} and \emph{How-level} instruction following. What-level instructions specify the task itself, while How-level instructions specify how to execute an already determined task, such as where to grasp an object or which path to take. Existing VLA research has largely focused on the What level, evaluating or improving robustness to paraphrased or replaced task instructions~\citep{zhou2025libero, fei2026libero, hou2026langgap, fang2026vision, kim2026libero, yang2026vision, lian2026langforce, fan2026interleave, zhan2026stable}. 
Meanwhile, systems requiring How-level guidance typically restrict it to predefined commands~\citep{shi2025hi, belkhale2024rt, chen2026steerable} or replace language corrections with visual symbols or externally generated recovery actions~\citep{zeng2026diagnose, lin2025failsafe}.
Consequently, whether VLAs can directly follow free-form How-level instructions, and what limits this ability, remains largely unexplored. We therefore ask: \textbf{Do current VLAs reflect How-level instructions in their actions, and if not, what prevents them from doing so?}

In this work, we systematically diagnose How-level instruction following in current VLAs through two complementary settings. First, in \emph{offline correction}, we collect failed episodes and rerun them with three alternatives: a simple retry with a different seed, a language correction generated from the observed failure, and an inverted correction that reverses the actionable content of the original correction. Comparing these alternatives allows us to determine whether a VLA uses the information in a correction rather than merely benefiting from another attempt. Second, in \emph{online intervention}, we examine a dual-system design~\citep{shi2025hi} in which a VLM monitors robot execution and provides corrections or sub-step instructions to the VLA during the task. Together, these settings evaluate instruction following under favorable and controlled conditions in offline correction and under deployment conditions in online intervention.

Across general VLAs~\citep{black2024pi_0, chen2025internvla, community2026starvla} and models designed to strengthen instruction following~\citep{yang2026vision, lian2026langforce}, both settings reveal a consistent limitation, as summarized in Fig.~\ref{fig:teaser}(a) and (b). In offline correction, corrections succeed less often than simple retries and do not clearly outperform inverted corrections. In online intervention, How-level instructions generally fail to improve task success, while frequent interventions substantially degrade performance. These findings suggest that current VLAs do not reliably use How-level instructions to guide their actions, even when designed to follow task instructions more robustly.

To understand this limitation, we investigate the structure of VLA training data. In standard robot datasets, each episode is paired with a single instruction shared across all timesteps~\citep{walke2023bridgedata}, and these instructions exhibit limited linguistic diversity~\citep{wanna2026limited}. Although the robot's actions change throughout an episode, the instruction typically describes only the task. Language may therefore serve primarily as a task identifier, without teaching the model how instructions should influence task execution. We hypothesize that this training structure limits How-level instruction following. An attention-level analysis provides indirect support: models assign different amounts of attention to corrections, yet their actions fail to reflect them regardless of the attention received, suggesting that the limitation is not specific to a single architectural component.

We test this hypothesis by introducing \emph{enriched instructions} that describe the robot's movements at different segments of an episode, while keeping the underlying demonstrations unchanged. Fig.~\ref{fig:teaser}(c) shows how these instructions differ from the original instruction shared across an episode. Across five VLA model families, enriched instructions maintain comparable task success while substantially improving responsiveness to How-level guidance. In particular, actions follow the directions specified by corrections in about 84\% of evaluated episodes. The original models and controls trained on the original instructions with the same training budget reach about 53\%, close to the 50\% chance level. However, this improvement does not consistently translate into higher task success. One possible explanation is that task success on standard benchmarks depends largely on the scene~\citep{lian2026langforce} and on chance~\citep{jang2026verifier}, and many failures can be recovered through retries alone. Thus, task success may not reliably reflect whether a model actually follows the additional instruction.

To measure How-level instruction following more directly, we introduce~\proposed, an \textbf{I}nstruction-\textbf{N}ecessity benchmark built on SIMPLER~\citep{li2024evaluating}, in which the conditions required for success are given only in How-level instructions. Its tasks require following a specified path around a transparent barrier, grasping a particular part of an object, or placing an object at an instructed location. None of the evaluated models, including those trained with enriched instructions, reliably follows these instructions, suggesting that the gained responsiveness falls short under demanding conditions and does not transfer to out-of-distribution (OOD) How-level instructions. How-level instruction following thus remains an open challenge for current VLAs. 

Our contributions are summarized as follows.
\begin{itemize}
    \item \textbf{Diagnosis of How-level instruction following.}
    We introduce offline correction and online intervention, revealing that current VLAs do not reliably follow How-level instructions even when designed for instruction following.
    \item \textbf{Training data hypothesis and enriched instructions.}
    We identify episode-level instruction supervision as a potential cause and show that enriched segment-level instructions improve action-level responsiveness without consistently improving task success.
    \item \textbf{\proposed.}
    We introduce an Instruction-Necessity benchmark that requires How-level instruction following for success, revealing remaining limitations of current VLAs.
\end{itemize}

\section{Related Work}
\label{sec:related_work}

\smallskip
\noindent\textbf{Instruction Following in VLAs.}
Recent studies have questioned whether VLAs genuinely follow language instructions or rely on visual cues and learned patterns.
LIBERO-PRO~\citep{zhou2025libero}, LIBERO-Plus~\citep{fei2026libero}, and LangGap~\citep{hou2026langgap} show that VLAs often exhibit limited behavioral changes when task instructions are modified or removed, while LIBERO-CF~\citep{fang2026vision} reveals their tendency to rely on visual priors even when different tasks share the same scene.
These findings are consistent with the limited linguistic diversity of robot datasets~\citep{wanna2026limited} and observations that standard SIMPLER~\citep{li2024evaluating} tasks can often be completed without language instructions~\citep{jang2026verifier}.
In contrast, LIBERO-Para~\citep{kim2026libero} reports that paraphrasing task instructions can substantially degrade performance in LIBERO-Goal~\citep{liu2023libero}, where multiple tasks share the same scene.
Together, these studies suggest that VLA behavior may depend on visual priors or familiar instruction patterns rather than the meaning of language instructions.
However, they primarily examine whether VLAs identify \emph{what} task to perform, leaving their ability to follow instructions specifying \emph{how} to perform a given task unexplored.

Another line of work makes VLAs follow task instructions more robustly. InstructVLA~\citep{yang2026vision} and ChatVLA~\citep{zhou2025chatvla} preserve the multimodal understanding of the VLM backbone. LangForce~\citep{lian2026langforce} and Residual Semantic Steering~\citep{zhan2026stable} reduce the policy's reliance on visual priors so that the task follows the instruction rather than the scene, even when the instruction is paraphrased. Spatially guided training~\citep{ye2026spatially} and interleaved image-text instructions~\citep{fan2026interleave} help the policy locate the object in the instruction. These methods aim to identify the intended task and object robustly from the instruction, and whether they also enable How-level instruction following remains unknown. Therefore, our diagnosis includes InstructVLA and LangForce.

\smallskip
\noindent\textbf{Correction and Hierarchical Control.}
Language has also been used to guide robots beyond the task instruction. In hierarchical control, an LLM or VLM plans the task, while the low-level policy executes only a fixed set of skills or commands learned during training, such as atomic subtasks, language motions, or structured steering commands~\citep{ahn2022can, huang2022inner, shi2025hi, belkhale2024rt, shi2024yell, chen2026steerable}. Training on such commands makes the low-level policy follow them, but it serves as an executor of predefined commands rather than as a general policy that both perform the task and interprets How-level language.
Failure correction and recovery methods likewise convert How-level corrections into visual symbols~\citep{zeng2026diagnose} or replace them with recovery actions generated by a separate VLM~\citep{lin2025failsafe}. In both lines of work, free-form How-level guidance bypasses the VLA. Whether a VLA can directly follow free-form How-level instructions, and why it fails to do so, has therefore not been diagnosed.

\section{Problem Formulation}
\label{sec:problem_formulation}

\smallskip
\noindent\textbf{VLA policy and training data.}
A VLA policy $\pi_\theta$ maps an observation $o_t$ and a language instruction $\ell$ to an action $a_t = \pi_\theta(o_t, \ell;\ \xi)$, where $a_t$ may be an action chunk and $\xi$ is the random seed that governs the stochasticity of the policy, such as the noise of a diffusion or flow-matching action head. The policy is trained on demonstrations $\mathcal{D} = \{\tau_i\}_{i=1}^{N}$, where each episode $\tau_i = \{(o_t, a_t, \ell_i)\}_{t=1}^{T_i}$ pairs every timestep with the same instruction $\ell_i$.

\smallskip
\noindent\textbf{What- and How-level instruction following.}
We decompose an instruction into $\ell = (\ell^{\text{what}}, \ell^{\text{how}})$, a conceptual split rather than a computational one. The What-level component $\ell^{\text{what}}$ specifies the task, namely which object to manipulate and what to do with it, and the How-level component $\ell^{\text{how}}$ covers everything else that can help perform the task, such as the path to take, where to grasp or place an object, or context about the scene. A How-level instruction is added on top of $\ell^{\text{what}}$ and provides additional information for performing the task when it is accurate. Given the observation and the task, the amount of this information is greater than zero,
\setlength{\abovedisplayskip}{4pt}\setlength{\belowdisplayskip}{4pt}
\begin{equation}
    I(A^\star_t;\ L^{\text{how}} \mid O_t,\ L^{\text{what}}) > 0,
    \label{eq:how_info}
\end{equation}
where $I$ denotes conditional mutual information, uppercase letters denote the random variables of the corresponding lowercase quantities, and $A^\star_t$ is an action that accomplishes the task. For a policy that uses this information, adding an accurate $\ell^{\text{how}}$ should raise task success, or at least not lower it.

\begin{table*}[!t]
    \centering
    \small
    \setlength{\tabcolsep}{3.5pt}
    \resizebox{\textwidth}{!}{%
    \begin{tabular}{l cccccc ccccc}
        \toprule
        & \multicolumn{6}{c}{Offline correction}
        & \multicolumn{5}{c}{Online intervention} \\
        \cmidrule(lr){2-7} \cmidrule(lr){8-12}
        Model & $N$ & Retry & Correction & $p$ (C vs.\ R)
        & Inverted & $p$ (C vs.\ I)
        & No intervention & Corrector & $\Delta$ & Unified & $\Delta$ \\
        \midrule
        \multicolumn{12}{l}{\textbf{\textit{General VLAs}}} \\
        $\pi_0$
        & 133 & 36.1 & 17.3 & \textbf{.001} & 11.3 & .185
        & 70.8 (2.8) & 72.9 (4.2) & \textbf{+2.1}
        & 58.0 (3.0) & $-$12.8 \\
        InternVLA-M1
        & 115 & 26.1 & 13.0 & \textbf{.028} & 18.3 & .307
        & 72.2 (0.9) & 70.5 (1.3) & $-$1.7
        & 58.0 (1.3) & $-$14.2 \\
        QwenGR00T
        & 142 & 22.5 & 8.5 & \textbf{<.001} & 7.0 & .815
        & 65.3 (0.9) & 61.1 (3.6) & $-$4.2
        & 53.5 (2.1) & $-$11.8 \\
        \textit{Pooled}
        & 390 & 28.2 & 12.8 & -- & 11.8 & --
        & 69.4 (1.0) & 68.2 (0.3) & $-$1.2
        & 56.5 (1.8) & $-$12.9 \\
        \midrule
        \multicolumn{12}{l}{\textbf{\textit{Instruction-following VLAs}}} \\
        InstructVLA
        & 210 & 11.9 & 7.1 & .099 & 4.3 & .286
        & 52.1 (2.8) & 49.7 (2.8) & $-$2.4
        & 34.7 (2.4) & $-$17.4 \\
        LangForce
        & 84 & 9.5 & 6.0 & .549 & 1.2 & .219
        & 73.6 (0.9) & 73.3 (1.9) & $-$0.3
        & 56.9 (3.1) & $-$16.7 \\
        \textit{Pooled}
        & 294 & 11.2 & 6.8 & -- & 3.4 & --
        & 62.8 (1.0) & 61.5 (2.4) & $-$1.3
        & 45.8 (0.5) & $-$17.0 \\
        \bottomrule
    \end{tabular}%
    }
    \caption{Diagnosis results for public VLA checkpoints (\%). \textbf{Offline correction:} $N$ is the number of failed episodes pooled over three seeds, and $p$ compares Correction (C) with Retry (R) and Inverted (I), with bold for $p<.05$. \textbf{Online intervention:} Entries show the mean over the three runs with standard errors in parentheses, and $\Delta$ is the change from No intervention, with bold for positive values. \textit{Pooled} rows aggregate the models in each group.}
    \label{tab:diagnosis}
\end{table*}

\section{Diagnosis: Do VLAs Follow How-Level Instructions?}
\label{sec:diagnosis}
We diagnose How-level instruction following in two complementary settings that append How-level instructions to the task instruction, examining the expectation in Eq.~\eqref{eq:how_info}. Offline correction examines it under favorable and controlled conditions, while online intervention examines it as in deployment. We evaluate general VLAs, $\pi_0$~\citep{black2024pi_0}, InternVLA-M1~\citep{chen2025internvla}, and QwenGR00T~\citep{community2026starvla}, and VLAs designed to strengthen instruction following, InstructVLA~\citep{yang2026vision} and LangForce~\citep{lian2026langforce}. Fig.~\ref{fig:teaser}(a) and (b) illustrate the procedure of each setting and the aggregated results over these five models. All evaluations follow the WidowX tasks of SIMPLER~\citep{li2024evaluating}.

\subsection{Offline Correction}
\label{sec:offline_correction}
Offline correction examines How-level instruction following under favorable and controlled conditions, in which each failure receives a verified correction and every other factor stays fixed. We first run each model with a random seed $\xi$ for up to 60 steps and collect the failed episodes. For each failure, we provide a proprietary VLM~\citep{gemini31pro2026} with the video of the failed episode, the task instruction, and a timeline of physical phases extracted from the simulator state, such as approaching, grasping, and transporting the object, together with the gripper-object distance and the end-effector height in each phase. The VLM first identifies what went wrong and when, and then writes a correction $c$ of one or two sentences specific to that failure. We refined the prompt by inspecting its outputs and manually checked 50 randomly sampled corrections, confirming that 44 are grounded in the actual failures.  The prompts, checking procedure, and details are in App.~\ref{app:correction}.

Each failed episode is then rerun from the same initial state under three conditions.
A \textit{retry} $(\ell;\ \xi')$ changes only the seed and adds no correction, so it shows how much the outcome changes by chance.
A \textit{correction} $(\ell \oplus c;\ \xi)$ appends $c$ to the task instruction and keeps the original seed.
An \textit{inverted correction} $(\ell \oplus \bar{c};\ \xi)$ appends $\bar{c}$ instead.
We obtain $\bar{c}$ by prompting the same VLM to reverse all actionable content of $c$, such as the direction, the object part to contact, the timing, and the gripper cue, while keeping its form.
An accurate correction provides task-relevant How-level information, as motivated by Eq.~\eqref{eq:how_info}, whereas an inverted correction provides misleading guidance.
If a policy effectively uses this information to improve task recovery, we would expect the idealized ordering \textit{correction} $>$ \textit{retry} $>$ \textit{inverted correction} in success rate.
In particular, comparing \textit{corrections} with \textit{inverted corrections} assesses whether the model benefits from the actionable content of the instruction, while comparing \textit{corrections} with \textit{retry} assesses whether the guidance improves recovery beyond stochastic variation.
This ordering is not a necessary condition for instruction following, as following an instruction does not always lead to successful task recovery.
For each condition, we report the rerun success rate, which is the fraction of failed episodes that succeed when rerun. Since each seed produces a different set of failed episodes that varies in number and difficulty, we pool the failed episodes collected with three seeds. We compare the \textit{correction} with the \textit{retry} and with the \textit{inverted correction} using two-sided exact McNemar tests~\citep{mcnemar1947note}, pairing the outcomes on the same failed episodes. We keep the original seed for the two corrections so that they act on the same behavior that failed, and App.~\ref{app:seed} reports corrections rerun with a new seed.

\subsection{Online Intervention}
\label{sec:online_intervention}
Online intervention evaluates How-level instruction following in a dual-system design, the setting in which a deployed VLA may receive How-level instructions~\citep{shi2025hi}.
A VLM serving as System~2 observes the rollout and guides the VLA serving as System~1.
Unlike offline correction, instructions arrive during execution and are generated from the observations available at each moment, allowing us to examine How-level instruction following under realistic conditions.

We use Qwen3-VL-8B~\citep{bai2025qwen3} as System~2 and compare three modes for System~1.
With \textit{no intervention}, the VLA executes the task using only the original task instruction $\ell$, without System~2.
With the \textit{corrector}, System~2 periodically monitors execution and generates a correction $c_t$ only when it detects a failure.
Here, $c_t$ is a free-form instruction describing how to recover from the detected failure, appended to the task instruction as $\ell \oplus c_t$.
Otherwise, the VLA continues with $\ell$ alone.
With the \textit{unified} mode, System~2 generates a sub-step instruction $s_t$ at every monitoring step, describing how to proceed with the next sub-step.
The VLA receives $\ell \oplus s_t$, where $s_t$ also provides corrective guidance when a failure is detected.
Here, $t$ denotes a monitoring step of System~2.

To leave room for these instructions to take effect, each episode runs for up to 240 steps, four times the 60 steps of the original SIMPLER evaluation.
We report the success rate under each mode as the mean and standard error over three independent runs on the same 96 episodes.\footnote{Standard errors in Table~\ref{tab:diagnosis} stay around the binomial noise of about three points, so the estimates are reliable.}
The prompts and detailed settings of System~2 are given in App.~\ref{app:online}.

\subsection{Results}
\label{sec:diagnosis_results}
Table~\ref{tab:diagnosis} shows that How-level instructions not only fail to help current VLAs in either setting but also lower their task success in most cases.

In offline correction, \textit{retry} recovers the most failed episodes for every model, so no model shows the expected ordering. All general VLAs recover clearly fewer episodes with \textit{corrections} than with \textit{retry}, and their \textit{corrections} do not clearly differ from \textit{inverted corrections}, with \textit{inverted corrections} recovering more for InternVLA-M1. The VLAs designed to strengthen instruction following show a similar pattern. Therefore, no model clearly benefits from the content of \textit{corrections}.

In online intervention, which reflects how How-level instructions would be given in deployment, these instructions do not raise task success and even lower it. The \textit{corrector} lowers task success for every model except $\pi_0$. Moreover, the \textit{unified} mode, which always intervenes, substantially lowers task success for every model. The drop is larger for the two VLAs designed to strengthen instruction following. Current VLAs thus do not benefit from the context that System~2 adds during execution, even when it targets a detected failure, and strengthening What-level instruction following does not make How-level instructions more useful.

Taken together, the results under both the favorable and controlled conditions of offline correction and the deployment conditions of online intervention suggest that \textit{How-level instructions act as noise rather than as context that helps current VLAs generate actions.} A similar pattern appears not only in the general VLAs but also in VLAs that strengthen instruction following at the What level. We next examine in Sec.~\ref{sec:data_hypothesis} where this failure, shared by both groups, may originate.

\section{Hypothesis: Data Structure Limits How-Level Instruction Following}
\label{sec:data_hypothesis}

\begin{table}[!htbp]
    \centering
    \scriptsize
    \setlength{\tabcolsep}{1pt}
    \begin{tabular*}{\linewidth}{@{\extracolsep{\fill}}lccccc@{}}
        \toprule
        Model & $\pi_0$ & InternVLA-M1 & QwenGR00T & InstructVLA & LangForce \\
        \midrule
        Ratio & 0.91 & 0.70 & 0.72 & 0.41 & 1.03 \\
        \bottomrule
    \end{tabular*}
    \caption{Ratio of the attention.}
    \label{tab:attention}
\end{table}

\begin{table*}[!t]
    \centering
    \small
    \setlength{\tabcolsep}{3.5pt}
    \begin{tabular}{l cccccc ccccc}
        \toprule
        & \multicolumn{6}{c}{Offline correction} & \multicolumn{5}{c}{Online intervention} \\
        \cmidrule(lr){2-7} \cmidrule(lr){8-12}
        Model & $N$ & Retry & Correction & $p$ & Inverted & $p$ & No intervention & Corrector & $\Delta$ & Unified & $\Delta$ \\
        \midrule
        \multicolumn{12}{l}{\textbf{\textit{Control}}} \\
        $\pi_0$         & 195 & 25.1 & 12.3 & \textbf{.001} & 6.7 & .080 & 63.5 (1.2) & 60.1 (2.5) & $-$3.4 & 45.5 (1.3) & $-$18.0 \\
        InternVLA-M1    & 218 & 5.5 & 2.8 & .146 & 2.8 & 1.000 & 52.4 (2.4) & 49.7 (2.1) & $-$2.7 & 42.0 (1.4) & $-$10.4 \\
        QwenGR00T       & 194 & 10.8 & 7.2 & .296 & 4.6 & .267 & 57.3 (2.4) & 55.6 (2.4) & $-$1.7 & 36.8 (1.7) & $-$20.5 \\
        InstructVLA     & 236 & 11.0 & 2.5 & \textbf{<.001} & 2.5 & 1.000 & 27.8 (3.6) & 26.7 (0.9) & $-$1.1 & 24.3 (0.7) & $-$3.5 \\
        LangForce       & 203 & 14.3 & 5.9 & \textbf{.009} & 3.0 & .180 & 53.5 (1.4) & 54.2 (1.0) & \textbf{+0.7} & 40.6 (1.2) & $-$12.9 \\
        \textit{Pooled} & 1046 & 13.1 & 5.9 & -- & 3.8 & -- & 50.9 (1.4) & 49.2 (0.7) & $-$1.7 & 37.8 (0.6) & $-$13.1 \\
        \midrule
        \multicolumn{12}{l}{\textbf{\textit{Enriched}}} \\
        $\pi_0$         & 193 & 21.8 & 11.9 & \textbf{.007} & 7.3 & .136 & 59.4 (1.6) & 56.6 (1.3) & $-$2.8 & 49.0 (0.6) & $-$10.4 \\
        InternVLA-M1    & 176 & 16.5 & 21.0 & .312 & 13.1 & \textbf{.016}$^\dagger$ & 63.9 (1.9) & 63.2 (0.3) & $-$0.7 & 56.2 (2.8) & $-$7.7 \\
        QwenGR00T       & 191 & 12.6 & 10.5 & .627 & 11.0 & 1.000 & 47.2 (3.3) & 49.0 (3.1) & \textbf{+1.8} & 51.0 (2.8) & \textbf{+3.8} \\
        InstructVLA     & 217 & 8.8 & 7.4 & .728 & 6.5 & .832 & 43.4 (0.9) & 45.1 (2.4) & \textbf{+1.7} & 44.4 (0.9) & \textbf{+1.0} \\
        LangForce       & 186 & 13.4 & 11.3 & .627 & 10.8 & 1.000 & 50.3 (2.7) & 49.3 (1.3) & $-$1.0 & 53.1 (2.6) & \textbf{+2.8} \\
        \textit{Pooled} & 963 & 14.4 & 12.1 & -- & 9.6 & -- & 52.8 (1.1) & 52.6 (0.3) & $-$0.2 & 50.8 (0.8) & $-$2.0 \\
        \bottomrule
    \end{tabular}
    \caption{Results of the enriched models and their controls, following the notation of Table~\ref{tab:diagnosis}. $^\dagger$ marks $p<.05$ with Correction higher, as expected from Eq.~\eqref{eq:how_info}.}
    \label{tab:enriched}
\end{table*}

We hypothesize that the failure in Sec.~\ref{sec:diagnosis} stems from the structure of training data, and test this hypothesis by changing only the training instructions.

\subsection{Hypothesis}
\label{sec:hypothesis}
As defined in Sec.~\ref{sec:problem_formulation}, each training episode pairs every timestep with the same instruction $\ell_i$.
In widely used robot datasets such as BridgeData~V2~\citep{walke2023bridgedata}, this instruction typically specifies the overall task.
Within an episode, the robot approaches, grasps, and transports an object while the instruction stays unchanged, so the model may predict actions primarily from the changing visual observations and use language mainly to identify the task.
This training structure provides little supervision for learning how different How-level instructions should influence actions for a given task.
\textit{We therefore hypothesize that episode-level instruction supervision contributes to VLAs' limited ability to follow How-level instructions.}

\smallskip
\noindent\textbf{Attention-Level Analysis.}
We further investigate whether the observed failure can be explained by insufficient attention to corrections.
Using failed episodes from the public models in offline correction, we append each correction to the original task instruction and examine the attention received by each at the component through which language information is passed to action generation.
We measure attention on frames from the portion of each episode where the failure occurs.
As shown in Table~\ref{tab:attention}, the ratio of attention received by the correction to that received by the task instruction varies across models, ranging from 0.41 to 1.03.
Nevertheless, corrections do not clearly outperform inverted corrections for any model, including those assigning comparable attention to corrections and task instructions.
These results suggest that the failure is not simply explained by insufficient attention to corrections.
Although attention alone cannot establish how language influences actions or rule out architectural factors, this observation motivates investigating the training data as a potential shared cause.
The measured components and detailed procedure are provided in App.~\ref{app:attention}.

\subsection{Enriched Instructions}
\label{sec:enriched_instructions}
If our hypothesis holds, a VLA trained on instructions that describe how the motion changes within an episode should respond to How-level instructions.
We therefore change only the training instructions, adding enriched instructions that describe \textit{how the robot moves in each segment of an episode.} We build them from the embodied chain-of-thought (ECoT) annotations~\citep{zawalski2024robotic} of BridgeData~V2~\citep{walke2023bridgedata}, which label every timestep with a subtask and a motion primitive. 
We use these public annotations rather than creating new ones, keeping our intervention minimal and focused on the structure of instructions within each episode.
We divide each episode into segments where the subtask changes and merge segments where the robot stays into a neighboring segment. For each segment, GPT-5.4-mini~\citep{openai2026gpt54mini} receives the task, the subtask, the motion primitives within it, such as moving left or closing the gripper, the gripper state, and the segment position in the episode. It then writes one imperative sentence that serves as a How-level instruction for the segment. 
The resulting data contain 272,662 enriched instructions for 43,771 episodes, and Fig.~\ref{fig:teaser}(c) shows examples.

We then train the five VLAs diagnosed in Sec.~\ref{sec:diagnosis} following the original training protocol of each model with LoRA~\citep{hu2022lora}, appending the enriched instruction of each segment to the task instruction. 
For a fair comparison, each model also has a control trained with the same protocol but only on the original task instructions, without enriched instructions. During evaluation, both the enriched models and the controls receive How-level instructions appended to the task instruction in the same way. The generation prompt, examples, training details, and ECoT results are in App.~\ref{app:training}.

\subsection{Results}
\label{sec:enriched_results}
Table~\ref{tab:enriched} reports the results of the enriched models and their controls under the same diagnosis as in Sec.~\ref{sec:diagnosis}. Without intervention, the enriched models reach task success comparable to the controls on average, so adding enriched instructions keeps the ability to perform the original tasks.

In online intervention, enriched instructions allow How-level instructions to raise task success, or at least reduce the harm of intervention. With the \textit{unified} mode, every enriched model either gains task success or loses less than the corresponding control. The \textit{corrector} shows the same pattern for most models with smaller changes.

Offline correction shows a similar change, as corrections recover more failed episodes for the enriched models than for the controls, while retries recover about as many. However, corrections exceed retries only for the enriched InternVLA-M1, and the advantage of corrections over inverted corrections or a neutral sentence in App.~\ref{app:neutral} does not appear consistently across models.

\begin{figure}[!htbp]
    \centering
    \includegraphics[width=0.85\linewidth]{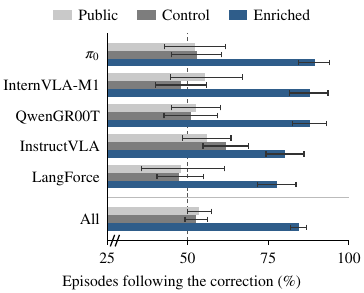}
    \caption{Action-level analysis. Error bars show 95\% confidence intervals, the line marks chance level, and \textit{All} combines the five model families.}
    \label{fig:action}
\end{figure}

Since the changes in How-level instruction following are inconsistent across models, we examine directly whether actions follow the content of a correction. To this end, we compare the trajectories under the correction and the inverted correction, which specify opposite directions for the same failed episode, such as moving the gripper up or down and to the left or right. If actions follow the content of these instructions, the trajectory under the correction should lie further in the specified direction than that under the inverted correction. We count an episode as following the instruction when this holds, so chance level is 50\%. As shown in Fig.~\ref{fig:action}, the public checkpoints and the controls stay near the chance level, whereas the enriched models consistently follow the specified direction in most episodes. \textit{Changing only the training instructions thus makes actions respond to How-level instructions}, which supports the hypothesis. App.~\ref{app:action} details the analysis and shows that the enriched models do not merely exceed the others but move strongly beyond them in the specified direction.

Taken together, enriched instructions make actions respond to How-level instructions in most episodes, but this response does not consistently lead to higher task success. One reason may be that task success reflects instruction following only indirectly, as the tasks can often be solved without relying on the instruction~\citep{lian2026langforce, jang2026verifier} and many failures are recovered by retries alone. We therefore evaluate tasks that can be solved only by following the instruction in Sec.~\ref{sec:benchmark}.

\section{\proposed: An Instruction-Necessity Benchmark}
\label{sec:benchmark}
As discussed in Sec.~\ref{sec:enriched_results}, task success on standard benchmarks reflects How-level instruction following only indirectly. To measure How-level instruction following more clearly, we build \proposed~on SIMPLER~\citep{li2024evaluating}, in which the How-level conditions required for task success exist only in the instruction.

\begin{figure}[!htbp]
    \centering
    \begingroup
    \footnotesize
    \setlength{\tabcolsep}{4pt}
    \begin{tabular}{@{}p{29.8pt}p{57pt}p{\dimexpr\linewidth-102.8pt\relax}@{}}
        \toprule
        \raggedright Task & \raggedright Condition & \raggedright Instruction \tabularnewline
        \midrule
        \raggedright Path & \raggedright Open side:\newline far / near & \raggedright ``there is a transparent barrier. approach the carrot from the \textbf{far~side}, \ldots'' \tabularnewline
        \raggedright Grasp & \raggedright Carrot end:\newline green / orange & \raggedright ``grasp the carrot by the \textbf{green~end}.'' \tabularnewline
        \raggedright Place & \raggedright Towel half:\newline far / near & \raggedright ``place it on the \textbf{far~part} of the towel, away from the robot.'' \tabularnewline
        \bottomrule
    \end{tabular}
    \endgroup
    \par\smallskip
    {\small (a) Task examples}
    \par\medskip
    \includegraphics[width=\linewidth]{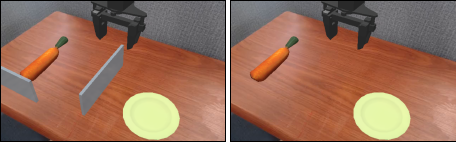}
    \par
    {\small (b) Path task scene}
    \caption{Overview of \proposed. (a) Task examples with their conditions and instructions. (b) The \textit{Path} scene with the barrier shown (left) and as seen (right).}
    \label{fig:benchmark}
\end{figure}

\subsection{Benchmark Design}
\label{sec:benchmark_design}
Each task in \proposed~has a How-level condition that determines the behavior required for success, such as the side to approach an object from. The condition is absent from the visual observation and is given only in the How-level instruction appended to the task instruction. The three tasks cover the \textit{path} to take, where to \textit{grasp} an object, and where to \textit{place} it, following the definition of How-level instructions in Sec.~\ref{sec:problem_formulation}, as listed in Fig.~\ref{fig:benchmark}(a).

In \textit{Path}, a transparent barrier blocks one side of the carrot, and the instruction specifies the open side to approach it from. As shown in Fig.~\ref{fig:benchmark}(b), the barrier is invisible in the observation since it is transparent, and an episode follows the instruction when the gripper reaches the carrot without passing over the barrier. The policy must therefore follow the instructed trajectory around an unseen barrier, which goes beyond the directions in Fig.~\ref{fig:action} and makes \textit{Path} a harder form of How-level instruction following. 
In \textit{Grasp}, the instruction indicates one end of the carrot, and an episode follows the instruction when the gripper grasps that end. In \textit{Place}, the instruction indicates the far or near side of the towel midline for the spoon, and an episode follows the instruction when the spoon lands on that side. App.~\ref{app:benchmark_design} gives the details.

We evaluate the public checkpoints, their controls, and the enriched models from Sec.~\ref{sec:data_hypothesis} without additional training. We report the task success rate over 24 episodes per task and the percentage of successful episodes that follow the instruction. The two values of each condition, such as far and near, are each assigned to half of the episodes, so a policy that ignores the instruction still follows it in 50\% of the successful episodes by chance.

\begin{table}[!htbp]
    \centering
    \small
    \setlength{\tabcolsep}{4pt}
    \begin{tabular}{lccc}
        \toprule
        Model & \textit{Path} & \textit{Grasp} & \textit{Place} \\
        \midrule
        \multicolumn{4}{l}{\textbf{\textit{Control}}} \\
        $\pi_0$         & 25.0 / 0.0 & 83.3 / 50.0 & 50.0 / 50.0 \\
        InternVLA-M1    & 0.0 / -- & 50.0 / 50.0 & 54.2 / 38.5 \\
        QwenGR00T       & 12.5 / 0.0 & 58.3 / 50.0 & 0.0 / -- \\
        InstructVLA     & 0.0 / -- & 12.5 / 66.7 & 8.3 / 50.0 \\
        LangForce       & 4.2 / 0.0 & 45.8 / 27.3 & 4.2 / 100.0 \\
        \textit{Pooled} & 8.3 / 0.0 & 50.0 / 46.7 & 23.3 / 46.4 \\
        \midrule
        \multicolumn{4}{l}{\textbf{\textit{Enriched}}} \\
        $\pi_0$         & 37.5 / 0.0 & 50.0 / 41.7 & 58.3 / 42.9 \\
        InternVLA-M1    & 4.2 / 0.0 & 58.3 / 28.6 & 29.2 / 42.9 \\
        QwenGR00T       & 45.8 / 0.0 & 50.0 / 50.0 & 29.2 / 57.1 \\
        InstructVLA     & 0.0 / -- & 16.7 / 25.0 & 25.0 / 33.3 \\
        LangForce       & 16.7 / 0.0 & 37.5 / 22.2 & 45.8 / 45.5 \\
        \textit{Pooled} & 20.8 / 0.0 & 42.5 / 35.3 & 37.5 / 44.4 \\
        \midrule
        \multicolumn{4}{l}{\textbf{\textit{Public checkpoints}}} \\
        \textit{Pooled} & 14.2 / 0.0 & 55.0 / 54.5 & 50.8 / 52.5 \\
        \bottomrule
    \end{tabular}
    \caption{Results on \proposed~(\%). Each entry is task success rate / instruction following rate among successful episodes. Table~\ref{tab:app_benchmark} reports each public model.}
    \label{tab:benchmark}
\end{table}

\subsection{Results and Analysis}
\label{sec:benchmark_results}
In Table~\ref{tab:benchmark}, every group struggles to follow How-level instructions.
In \textit{Path}, no model follows the instruction, as every successful episode passes over the barrier, including those of the enriched models. 
This result suggests that the responsiveness of the enriched models remains insufficient for tasks that demand strict adherence to How-level instructions.

\begin{figure}[!htbp]
    \centering
    \includegraphics[width=0.85\linewidth]{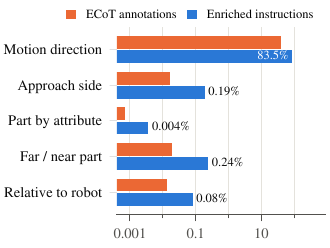}
    \caption{Frequency of How-level content in the ECoT annotations and the enriched instructions (log scale).}
    \label{fig:ecot_data}
\end{figure}

In \textit{Grasp} and \textit{Place}, the enriched models follow the instruction near the chance level as the controls and the public checkpoints do, which appears inconsistent with the hypothesis since they respond to How-level instructions in Sec.~\ref{sec:enriched_results}.
To investigate this inconsistency, we analyze the How-level content of the enriched instructions and the ECoT~\citep{zawalski2024robotic} annotations from which they are derived, by counting the sentences that contain each kind of content. As shown in Fig.~\ref{fig:ecot_data}, most of them specify the direction of motion.
In contrast, the approach side in \textit{Path}, the object part specified by an attribute in \textit{Grasp}, and the far or near part of an object in \textit{Place} rarely appear in either source. The instructions in these tasks thus lie outside the training distribution, which suggests that \textit{the responsiveness gained from enriched instructions does not extend to OOD How-level instructions.} App.~\ref{app:data_analysis} details the analysis.

Taken together, the enriched instructions partly support the hypothesis. As shown in Sec.~\ref{sec:enriched_results}, they make actions respond to How-level instructions and reduce the harm of intervention, yet this responsiveness does not consistently lead to higher task success. \proposed~further shows that the responsiveness remains limited to the How-level content seen during training and falls short in \textit{Path}, which also demands strict adherence.
Therefore, How-level instruction following remains an open challenge for current VLAs, and addressing it may require training data that covers How-level content more broadly. \proposed~provides a testbed for measuring progress on this challenge.

\section{Conclusion}
\label{sec:conclusion}
We find that current VLAs struggle to follow How-level instructions. Enriched instructions based on our data hypothesis make actions respond to them without consistently raising task success. On \proposed, which requires How-level instructions for task success, all models fail on demanding How-level instructions and do not follow OOD ones. We hope that our diagnoses and \proposed~help future work toward VLAs that follow How-level instructions.

\clearpage

\section*{Limitations}

Our experiments cover a single combination of embodiment, simulator, and training data, the WidowX robot in SIMPLER with BridgeData~V2, on short-horizon pick-and-place tasks. Other embodiments and longer tasks may behave differently, which we leave for future work. We manually check sampled corrections in offline correction, and online intervention assumes imperfect VLM-generated instructions by design. In online intervention, the enriched models gain task success in several cases under the same System~2, which indirectly suggests that the given instructions can be correct. Nevertheless, it remains possible that incorrect instructions account for part of the observed failures. Enriched instructions change both the temporal alignment of instructions and their How-level content. In the short-horizon tasks, the task is the same throughout an episode, so instructions that change across segments naturally describe how to carry out the task rather than what the task is. Segment-level instructions thus tend to be How-level, and our hypothesis treats them jointly. However, we acknowledge that our experiments do not separate them, and leave separating their contributions to future work. We train the enriched models with LoRA under a limited training budget, so the results may differ under full fine-tuning. Lastly, we use a transparent barrier in \textit{Path} to evaluate demanding How-level instruction following. Since the barrier is invisible in the observation, whereas transparent objects are often partly visible in the real world, results on \textit{Path} should be read as an evaluation of How-level instruction following rather than of realistic manipulation.

\bibliography{references}

\clearpage
\appendix
\section{Details of the Diagnosis}
\label{app:diagnosis}

\subsection{Generation of Corrections}
\label{app:correction}

\noindent\textbf{Failed episodes and timeline.}
Each model runs the four WidowX tasks of SIMPLER with 24 episodes each for up to 60 steps under the seeds 42, 142, and 242.
For each failure, we label every step with one of nine phases, such as approach, lift, and transport, from the grasp state, the distance from the end-effector to the object, the distance from the object to the target, and the end-effector height. The timeline lists each segment of consecutive steps with the same phase, together with the step range and the start and end values of these quantities.

\smallskip
\noindent\textbf{Prompt design and corrections.}
We design the prompt on 12 failed episodes. We compare three timeline formats, the raw phase segments, segments without short grasps, and segments with short ones merged, and three prompts that specify the critical point of the failure as a frame range, a window of the video in percent, or indices of timeline segments. By visually inspecting the 108 outputs, we select the raw segments and the prompt with segment indices, which give the most accurate corrections. As shown in Fig.~\ref{fig:prompt_correction}, Gemini~3.1 Pro then receives the video of the failed episode, the task instruction, and the timeline, reasons about what went wrong and when, selects the critical segments, and writes one correction of one or two sentences. A separate call to the same model, shown in Fig.~\ref{fig:prompt_inverted}, inverts all actionable content of the correction, including every direction, the object part to contact, the timing, and any gripper cue, while keeping the same object and destination. For example, ``Move the gripper further to the right to properly align with the spoon handle before closing it.'' becomes ``Move the gripper further to the left to properly align with the spoon bowl after opening it.''

\begin{figure*}[t]
    \centering
    \includegraphics[width=\textwidth]{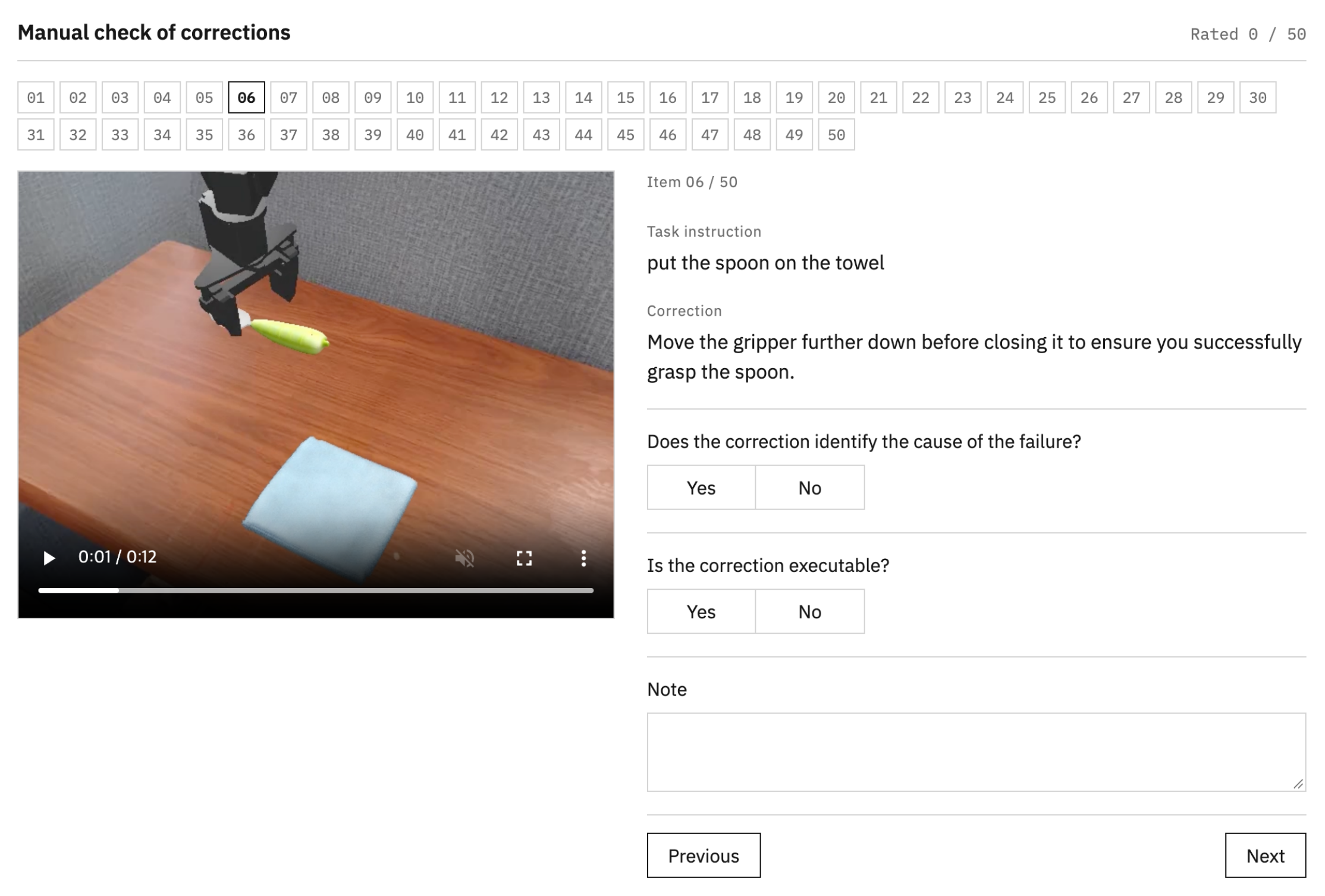}
    \caption{Interface for the manual check of corrections.}
    \label{fig:correction_check}
\end{figure*}

\smallskip
\noindent\textbf{Human verification.}
We randomly sample 50 failed episodes, ten from each model family across the public checkpoints, the controls, and the enriched models, excluding those used for prompt design. The authors watch each episode without knowing the model and check whether the correction identifies the cause of the failure and whether it can be executed, using the interface in Fig.~\ref{fig:correction_check}. As shown in Table~\ref{tab:app_check}, the corrections identify the cause in 44 of the 50 episodes (88\%), and all of them can be executed.

\begin{table}[!htbp]
    \centering
    \small
    \begin{tabular}{lccc}
        \toprule
        Group & $N$ & Cause & Executable \\
        \midrule
        Public   & 17 & 15 & 17 \\
        Control  & 17 & 13 & 17 \\
        Enriched & 16 & 16 & 16 \\
        \midrule
        All      & 50 & 44 & 50 \\
        \bottomrule
    \end{tabular}
    \caption{Number of corrections that identify the cause of the failure and that can be executed in the manual check.}
    \label{tab:app_check}
\end{table}

\smallskip
\noindent\textbf{Timing of corrections.}
The VLM writes each correction for the critical segments of the failure, whereas offline correction appends the correction to the task instruction before execution. We make this choice for two reasons. First, offline correction reflects a user who gives How-level advice together with the task before the robot starts. The corrections fit this setting, as they are written as How-level instructions that can be given in advance and about 90\% of them specify when to act with a temporal cue, such as ``Lower the gripper further down \textit{before closing it.}'' Second, instructions given during execution are covered by online intervention (Sec.~\ref{sec:online_intervention}).

We then examine whether this choice affects the results. For nearly half of the failures in each group (40 to 47\%), the critical segments start at the first step, so the correction already precedes them as the prompt intends. If giving corrections before execution made them fall short of retries, the shortfall would shrink for these failures. Table~\ref{tab:app_timing} shows that it does not. In both subsets, corrections recover fewer failed episodes than retries for the public checkpoints and the controls and about as many for the enriched models. The timing of corrections therefore does not explain why corrections fall short of retries.

\begin{table}[!htbp]
    \centering
    \small
    \setlength{\tabcolsep}{4pt}
    \begin{tabular}{llrccc}
        \toprule
        Group & Start & $N$ & R & C & I \\
        \midrule
        Public   & First & 274 & 19.0 & 7.3  & 6.9  \\
                 & Later & 410 & 22.2 & 12.2 & 9.0  \\
        \midrule
        Control  & First & 458 & 14.0 & 4.4  & 2.4  \\
                 & Later & 588 & 12.4 & 7.1  & 4.9  \\
        \midrule
        Enriched & First & 452 & 12.2 & 8.8  & 6.4  \\
                 & Later & 511 & 16.4 & 15.1 & 12.3 \\
        \bottomrule
    \end{tabular}
    \caption{Rerun success rate (\%) by whether the critical segments start at the first step, in the notation of Table~\ref{tab:diagnosis}.}
    \label{tab:app_timing}
\end{table}

\subsection{Corrections with the Retry Seed}
\label{app:seed}

Each correction is written for the failure observed under the original seed $\xi$.
We therefore keep $\xi$ for the two corrections so that they act on the same behavior that failed, since a new seed may lead to a different behavior that the correction does not describe.
Nevertheless, since the retry uses a new seed $\xi'$, one may ask whether part of the advantage of retries comes from the new seed rather than from the absence of a correction.
To clarify this, we rerun the corrections and inverted corrections with $\xi'$ on the same failed episodes, so that they share the seed with the retry and only the instruction changes.
Table~\ref{tab:app_seed} reports the results of all 15 models, pooled over the models in each group.
\begin{table}[!htbp]
    \centering
    \small
    \setlength{\tabcolsep}{4pt}
    \begin{tabular}{lrccccc}
        \toprule
         & & Retry & \multicolumn{2}{c}{Seed $\xi$} & \multicolumn{2}{c}{Seed $\xi'$} \\
        \cmidrule(lr){4-5} \cmidrule(lr){6-7}
        Group & $N$ & ($\xi'$) & C & I & C & I \\
        \midrule
        Public   & 684  & 20.9 & 10.2 & 8.2 & 10.8 & 10.5 \\
        Control  & 1046 & 13.1 & 5.9  & 3.8 & 6.6  & 5.3 \\
        Enriched & 963  & 14.4 & 12.1 & 9.6 & 11.4 & 10.5 \\
        \bottomrule
    \end{tabular}
    \caption{Rerun success rate (\%) under each seed, in the notation of Table~\ref{tab:diagnosis}.}
    \label{tab:app_seed}
\end{table}

With $\xi'$, corrections and inverted corrections show a pattern similar to that described in Secs.~\ref{sec:diagnosis_results} and~\ref{sec:enriched_results}.

\subsection{Online Intervention Settings}
\label{app:online}

\noindent\textbf{Monitoring.}
System~2 monitors execution every eight steps, which gives 30 monitoring steps in an episode of 240 steps. At each monitoring step, System~2 receives the task instruction and the frames at all monitoring steps so far, with the current frame last, and Qwen3-VL-8B-Instruct generates the output with greedy decoding.

\smallskip
\noindent\textbf{Modes and runs.}
Fig.~\ref{fig:prompt_online} shows the prompts. The \textit{corrector} outputs SKIP unless System~2 clearly sees a failure. A correction stays appended to the task instruction until the next monitoring step, and after SKIP nothing is appended, so the VLA receives the task instruction alone. The \textit{unified} mode outputs one instruction for the next motion at every monitoring step, which replaces the previous one. The three runs share the 24 scenes of each task and differ in the sampling noise of the policy.

\section{Details of the Data Hypothesis}
\label{app:hypothesis}

\begin{table*}[t]
    \centering
    \small
    \setlength{\tabcolsep}{2.5pt}
    \begin{tabular}{llllcc}
        \toprule
        Model & Initialization & Trained parameters & Control & Enriched & Budget \\
        \midrule
        $\pi_0$         & PaliGemma-3B                 & LoRA (r=32) on LM, full ViT and action expert & 48k & 51.2k & 100k \\
        InternVLA-M1    & InternVLA-M1 (stage 1)       & LoRA (r=32) on LLM, full Q-Former and action head        & 60k & 100k & 120k \\
        QwenGR00T       & Qwen3-VL-4B                  & LoRA (r=32) on VLM, action head                          & 40k & 40k & 50k \\
        InstructVLA     & Eagle2-2B (expert stage)     & LoRA (r=128) on LLM, action expert                       & 40k & 95k & 100k \\
        LangForce       & Qwen3-VL-4B                  & LoRA (r=32) on VLM, action head, action queries          & 40k & 40k & 50k \\
        \bottomrule
    \end{tabular}
    \caption{Training of the enriched models and their controls, which share the recipe. The Control and Enriched columns give the steps of the checkpoints with the highest task success in SIMPLER. Budget gives the total training steps.}
    \label{tab:app_training}
\end{table*}

\subsection{Attention-Level Analysis}
\label{app:attention}

\begin{table}[!htbp]
    \centering
    \small
    \setlength{\tabcolsep}{4pt}
    \begin{tabular}{lll}
        \toprule
        Model & Attention & Query \\
        \midrule
        $\pi_0$         & VLM, 18 layers        & All prefix \\
        InternVLA-M1    & Q-Former cross-attn.  & 64 queries \\
        QwenGR00T       & VLM, 36 layers        & All \\
        InstructVLA     & VLM, 6 layers         & 64 meta \\
        LangForce       & VLM, 36 layers        & 64 action \\
        \bottomrule
    \end{tabular}
    \caption{Where the attention in Table~\ref{tab:attention} is measured, averaged over heads, query tokens, and layers.}
    \label{tab:app_attention}
\end{table}

Table~\ref{tab:app_attention} lists where we measure attention in each model. We choose the component through which each model passes language to action generation. For InternVLA-M1, this is the cross-attention of the Q-Former that reads the last VLM layer, and for InstructVLA, it is the attention of the meta tokens that condition the action expert. For LangForce, it is the attention of the action queries whose outputs condition the action head. For $\pi_0$ and QwenGR00T, we average the attention that each token receives over all query positions of the VLM. We sample frames every two steps from the part of each failed episode in which the failure occurs, using failures from the first seed of offline correction, and pair each frame with the task instruction followed by the correction. This yields 685, 333, 504, 774, and 346 frames for $\pi_0$, InternVLA-M1, QwenGR00T, InstructVLA, and LangForce. The ratio counts only the tokens that overlap the text of the task instruction or of the correction, and excludes prompt-template words, the period that separates the two, special tokens such as beginning-of-sequence and role tokens, and image tokens. For each part, we divide the attention by the number of tokens and average the result over frames before taking the ratio.

\subsection{Enriched Instructions and Training}
\label{app:training}

\noindent\textbf{Enriched instructions.}
The enriched instructions are generated with the Batch API of GPT-5.4-mini at a temperature of 0.7, using text inputs only. For each segment, the input contains the task, the original instruction, the plan and subtask with their reasoning, the motion primitives with their reasoning, the gripper state and whether the state changes, and the position of the segment within the episode. As shown in Fig.~\ref{fig:prompt_enriched}, the prompt asks for one imperative sentence of about 25 to 30 words in the present tense, grounded in the motion primitives, with directions given relative to objects and without numerical quantities. For the first segment of an episode with the instruction ``put small spoon from basket to tray'', the generated sentence is ``Move down over the spoon in the basket until the gripper aligns at grasping height and closes around it.'' Of the 53,192 training episodes, 9,421 receive no enriched instruction and keep the original instruction alone. Table~\ref{tab:app_training} summarizes the training of each model.

\smallskip
\noindent\textbf{Checkpoint selection.}
For each model, we evaluate checkpoints without intervention on the four WidowX tasks of SIMPLER and select the one with the highest task success. The How-level evaluations are not used for selection.

\smallskip
\noindent\textbf{ECoT.}
Since the enriched instructions are built from the ECoT annotations, a natural question is whether the public ECoT checkpoint~\citep{zawalski2024robotic}, which learns to generate these annotations as reasoning before each action, also responds to How-level instructions.
To examine this, we apply offline correction to ECoT with seed 42, using sampling for the retry since ECoT decodes greedily.
ECoT succeeds in only 3 of the 96 episodes on the four WidowX tasks of SIMPLER, and rerunning the 93 failed episodes recovers 4 with retry, 3 with corrections, and none with inverted corrections.
These numbers are too small for comparison, so we do not include ECoT in our experiments.

\subsection{Appending a Neutral Sentence}
\label{app:neutral}

Corrections append a sentence to the task instruction, so their effect may come from the appended sentence rather than from the How-level content. The enriched models are also trained with sentences appended to the task instruction, so their recovery with corrections may come from this familiar format. To examine both, we rerun the failed episodes of all groups with $\xi$ while appending a neutral sentence, ``Please follow the task instruction as written and make sure to succeed at this task.'' The sentence has 15 words, close to the median length of the corrections.

\begin{table}[!htbp]
    \centering
    \small
    \setlength{\tabcolsep}{4pt}
    \begin{tabular}{lrcccc}
        \toprule
        Group & $N$ & R & C & I & Neutral \\
        \midrule
        Public   & 684  & 20.9 & 10.2 & 8.2  & 19.7 \\
        \quad $\pi_0$      & 133 & 36.1 & 17.3 & 11.3 & 31.6 \\
        \quad InternVLA-M1 & 115 & 26.1 & 13.0 & 18.3 & 35.7 \\
        \quad QwenGR00T    & 142 & 22.5 & 8.5  & 7.0  & 13.4 \\
        \quad InstructVLA  & 210 & 11.9 & 7.1  & 4.3  & 10.0 \\
        \quad LangForce    & 84  & 9.5  & 6.0  & 1.2  & 14.3 \\
        \midrule
        Control  & 1046 & 13.1 & 5.9  & 3.8  & 6.3  \\
        \quad $\pi_0$      & 195 & 25.1 & 12.3 & 6.7  & 3.6  \\
        \quad InternVLA-M1 & 218 & 5.5  & 2.8  & 2.8  & 7.3  \\
        \quad QwenGR00T    & 194 & 10.8 & 7.2  & 4.6  & 9.3  \\
        \quad InstructVLA  & 236 & 11.0 & 2.5  & 2.5  & 4.2  \\
        \quad LangForce    & 203 & 14.3 & 5.9  & 3.0  & 7.4  \\
        \midrule
        Enriched & 963  & 14.4 & 12.1 & 9.6  & 11.8 \\
        \quad $\pi_0$      & 193 & 21.8 & 11.9 & 7.3  & 7.3  \\
        \quad InternVLA-M1 & 176 & 16.5 & 21.0 & 13.1 & 15.9 \\
        \quad QwenGR00T    & 191 & 12.6 & 10.5 & 11.0 & 12.0 \\
        \quad InstructVLA  & 217 & 8.8  & 7.4  & 6.5  & 14.3 \\
        \quad LangForce    & 186 & 13.4 & 11.3 & 10.8 & 9.7  \\
        \bottomrule
    \end{tabular}
    \caption{Rerun success rate (\%) with the neutral sentence, in the notation of Table~\ref{tab:diagnosis}.}
    \label{tab:app_neutral}
\end{table}

Table~\ref{tab:app_neutral} reports the results. For the public checkpoints, the neutral sentence recovers more failed episodes than corrections and inverted corrections in every model. Appending a sentence therefore does not explain the drop, which comes from the How-level content acting as noise rather than as guidance, as found in Sec.~\ref{sec:diagnosis_results}. For the controls, the neutral sentence tends to recover fewer failed episodes than retries.

Within the enriched models, if the familiar format alone drove the recovery, the neutral sentence would recover about as many failed episodes as corrections. Instead, the neutral sentence tends to recover more failed episodes than corrections for QwenGR00T and InstructVLA, as retries do, and fewer than corrections for $\pi_0$, InternVLA-M1, and LangForce, as inverted corrections do. The enriched models thus respond to the content of the appended sentence rather than to the format, in line with the action-level analysis in Sec.~\ref{sec:enriched_results}, where actions follow the directions specified by corrections. However, since the neutral sentence recovers more failed episodes than corrections for some models, this response does not consistently lead to higher task success.

\subsection{Action-Level Analysis by Model}
\label{app:action}

\begin{table}[!htbp]
    \centering
    \small
    \begin{tabular}{lccc}
        \toprule
        Model & Public & Control & Enriched \\
        \midrule
        $\pi_0$         & 52.3 (109) & 52.9 (174) & 89.3 (169) \\
        InternVLA-M1    & 55.4 (101) & 48.0 (179) & 87.9 (149) \\
        QwenGR00T       & 52.6 (114) & 50.9 (159) & 88.1 (159) \\
        InstructVLA     & 56.0 (182) & 62.0 (213) & 80.2 (197) \\
        LangForce       & 48.0 (75)  & 47.4 (171) & 77.7 (166) \\
        \midrule
        All             & 53.5 (581) & 52.7 (896) & 84.4 (840) \\
        \bottomrule
    \end{tabular}
    \caption{Fraction (\%) of failed episodes that follow the directions of the correction, with the number of episodes in parentheses. The bootstrap 95\% confidence intervals of \textit{All} are [49.8, 57.4], [49.2, 56.1], and [81.8, 86.9].}
    \label{tab:app_action}
\end{table}

We label each pair of a correction and its inverted correction automatically before reading the trajectories. We detect words for the lateral directions, left and right, and for the vertical directions, such as lower, down, raise, and lift, and remove non-directional uses such as ``pick up''. We keep a pair only if every axis it mentions is clearly reversed, which leaves 2,317 of the 2,693 failed episodes of the three groups. For the lateral axis, we use the component of the end-effector position along the rightward direction of the camera, and for the vertical axis, we use the end-effector height. We average the position difference between the two trajectories over their common length and count an axis as correct when the difference has the sign specified by the correction. An episode follows the instruction when every specified axis is correct. Most pairs specify a single axis (536 of 581, 838 of 896, and 791 of 840 for the public checkpoints, the controls, and the enriched models), so the chance level under balanced and independent directions is about 48\%, close to 50\%. Table~\ref{tab:app_action} reports the result for each model. Over the five families, the lateral axis is correct in 60.6\%, 62.3\%, and 80.1\% of the cases for the public checkpoints, the controls, and the enriched models, and the vertical axis in 52.6\%, 50.8\%, and 87.1\%. Forward and backward directions, object parts, and timing are not evaluated, and the inverted correction may change them as well, so this analysis measures whether actions follow the specified directions rather than the whole correction.

\begin{table}[!htbp]
    \centering
    \small
    \setlength{\tabcolsep}{4pt}
    \begin{tabular}{lccc}
        \toprule
        Model & Public & Control & Enriched \\
        \midrule
        $\pi_0$         & 0.25 / 31.2 & 0.21 / 28.7 & 4.33 / 79.3 \\
        InternVLA-M1    & 0.18 / 29.7 & 0.03 / 24.6 & 1.05 / 52.3 \\
        QwenGR00T       & 0.18 / 30.7 & 0.11 / 32.7 & 2.09 / 62.9 \\
        InstructVLA     & 0.32 / 35.2 & 0.34 / 19.2 & 1.39 / 57.4 \\
        LangForce       & 0.09 / 34.7 & $-$0.01 / 25.1 & 1.60 / 57.2 \\
        \midrule
        All             & 0.21 / 32.5 & 0.15 / 25.7 & 1.74 / 61.9 \\
        \bottomrule
    \end{tabular}
    \caption{Median displacement (cm) in the specified direction / fraction (\%) of episodes that move more than 1\,cm in every specified direction.}
    \label{tab:app_displacement}
\end{table}

To check that the result does not rely on small differences, Table~\ref{tab:app_displacement} reports the size of the displacement. The enriched models move 1.74\,cm further in the specified direction in the median, compared with 0.21\,cm and 0.15\,cm for the public checkpoints and the controls, and the difference remains when we count only displacements larger than 1\,cm.

\section{Details of \proposed}
\label{app:simpler_in}

\subsection{Task Design and Judgment}
\label{app:benchmark_design}

\noindent\textbf{Common setup.}
\textit{Path} and \textit{Grasp} build on the carrot-on-plate task of SIMPLER, and \textit{Place} builds on the spoon-on-towel task. Each task has 24 episodes that follow the object layouts of SIMPLER, and each value of the condition is assigned to 12 episodes. Table~\ref{tab:app_instructions} lists the instructions. Each episode runs for up to 240 steps, and a scripted oracle confirms that each task can be solved.

\smallskip
\noindent\textbf{Judgment.}
In \textit{Path}, the barrier consists of two transparent panels 6\,cm tall that collide with the robot and the objects but are not rendered, one between the carrot and the plate and one closing the far or near side of the carrot. An episode follows the instruction when the task succeeds and the gripper never passes over either panel. 
In \textit{Grasp}, we use a carrot of 15.4\,cm whose two ends can both be grasped, and judge the grasped end by projecting the contact point onto the long axis of the carrot once the grasp becomes stable. In \textit{Place}, we enlarge the towel by 1.75 times and judge the landing side by comparing the spoon center with the towel center along the far--near direction when the spoon first touches the towel.

\begin{table}[!htbp]
    \centering
    \small
    \setlength{\tabcolsep}{3pt}
    \begin{tabular}{@{}lp{0.78\linewidth}@{}}
        \toprule
        Task & Instruction \\
        \midrule
        \textit{Path}  & ``put carrot on plate. there is a transparent barrier. approach the carrot from the \textbf{far side, behind the carrot, away from the robot}, and exit the same way.'' \\
                       & ``put carrot on plate. there is a transparent barrier. approach the carrot from the \textbf{near side, in front of the carrot, close to the robot}, and exit the same way.'' \\
        \midrule
        \textit{Grasp} & ``put carrot on plate. grasp the carrot by the \textbf{green end}.'' \\
                       & ``put carrot on plate. grasp the carrot by the \textbf{orange end}.'' \\
        \midrule
        \textit{Place} & ``put the spoon on the towel. place it on the \textbf{far part} of the towel, away from the robot.'' \\
                       & ``put the spoon on the towel. place it on the \textbf{near part} of the towel, close to the robot.'' \\
        \bottomrule
    \end{tabular}
    \caption{Instructions in \proposed, each consisting of the task instruction and the How-level instruction, with the condition in bold.}
    \label{tab:app_instructions}
\end{table}

\begin{table}[!htbp]
    \centering
    \small
    \setlength{\tabcolsep}{4pt}
    \begin{tabular}{lccc}
        \toprule
        Model & \textit{Path} & \textit{Grasp} & \textit{Place} \\
        \midrule
        $\pi_0$         & 33.3 / 0.0 & 70.8 / 47.1 & 83.3 / 60.0 \\
        InternVLA-M1    & 20.8 / 0.0 & 75.0 / 55.6 & 37.5 / 44.4 \\
        QwenGR00T       & 12.5 / 0.0 & 41.7 / 50.0 & 58.3 / 57.1 \\
        InstructVLA     & 4.2 / 0.0 & 25.0 / 50.0 & 16.7 / 50.0 \\
        LangForce       & 0.0 / -- & 62.5 / 66.7 & 58.3 / 42.9 \\
        \textit{Pooled} & 14.2 / 0.0 & 55.0 / 54.5 & 50.8 / 52.5 \\
        \bottomrule
    \end{tabular}
    \caption{Results of each public checkpoint on \proposed, in the notation of Table~\ref{tab:benchmark}.}
    \label{tab:app_benchmark}
\end{table}

\subsection{Analysis of How-Level Content}
\label{app:data_analysis}

\begin{table}[!htbp]
    \centering
    \small
    \setlength{\tabcolsep}{3pt}
    \begin{tabular}{@{}lp{0.36\linewidth}rr@{}}
        \toprule
        Content & Pattern & ECoT & Enriched \\
        \midrule
        Motion direction  & left, right, up, down, upward, downward, forward, backward, lower, lift, raise & 40.3 & 83.5 \\
        Approach side     & from the left / right / front / back / far / near / top / other side / end & 0.016 & 0.19 \\
        Part by attribute & color word + end / tip / part / side / half & 0.0007 & 0.004 \\
        Far / near part   & far / near + part / side / half / end / edge / corner & 0.019 & 0.24 \\
        Relative to robot & away from / toward(s) / close(r) to + (the) robot / you / camera / base & 0.014 & 0.08 \\
        \bottomrule
    \end{tabular}
    \caption{Patterns and frequencies (\%) of the How-level content in Fig.~\ref{fig:ecot_data}. Color words are green, orange, red, blue, yellow, white, black, brown, purple, and pink.}
    \label{tab:app_content}
\end{table}

We measure how often each kind of How-level content appears in two sources, the subtask annotations of ECoT and the enriched instructions. For ECoT, we keep each distinct subtask annotation once per episode, which gives 280,526 sentences, and for the enriched instructions, we take all 272,662 sentences. We lowercase each sentence and count the sentence for a kind of content when the sentence contains a whole-word match to the pattern of that kind in Table~\ref{tab:app_content}, so one sentence may count for several kinds. The frequency of each kind is the share of such sentences in each source. The other ECoT annotations, such as the plan, show similar frequencies.

\section{Prompts}
\label{app:prompts}

\begin{figure*}[p]
    \centering
    \includegraphics[width=\textwidth]{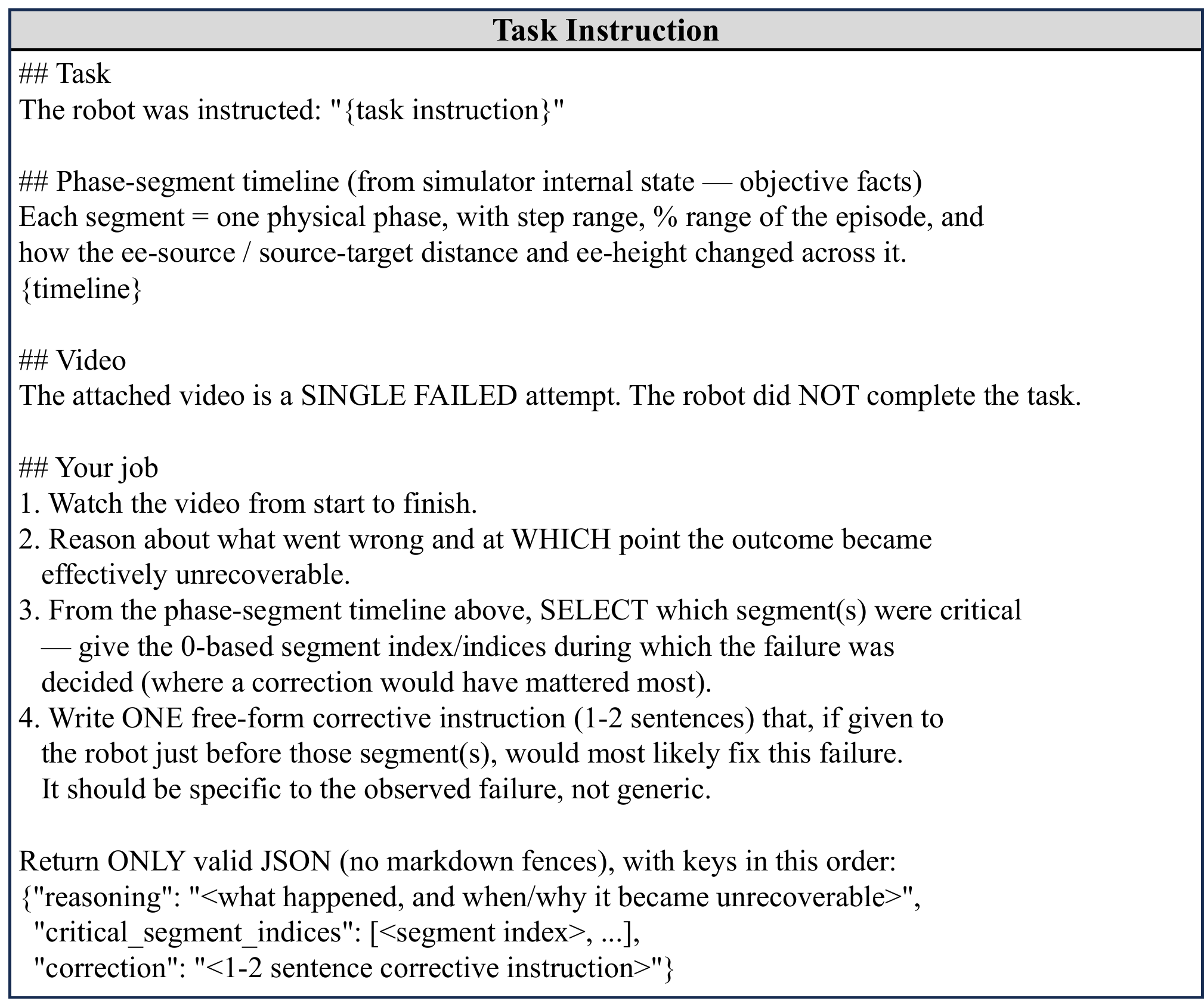}
    \caption{Prompt for generating corrections with Gemini~3.1 Pro. Placeholders in braces are filled for each failed episode.}
    \label{fig:prompt_correction}
\end{figure*}

\begin{figure*}[p]
    \centering
    \includegraphics[width=\textwidth]{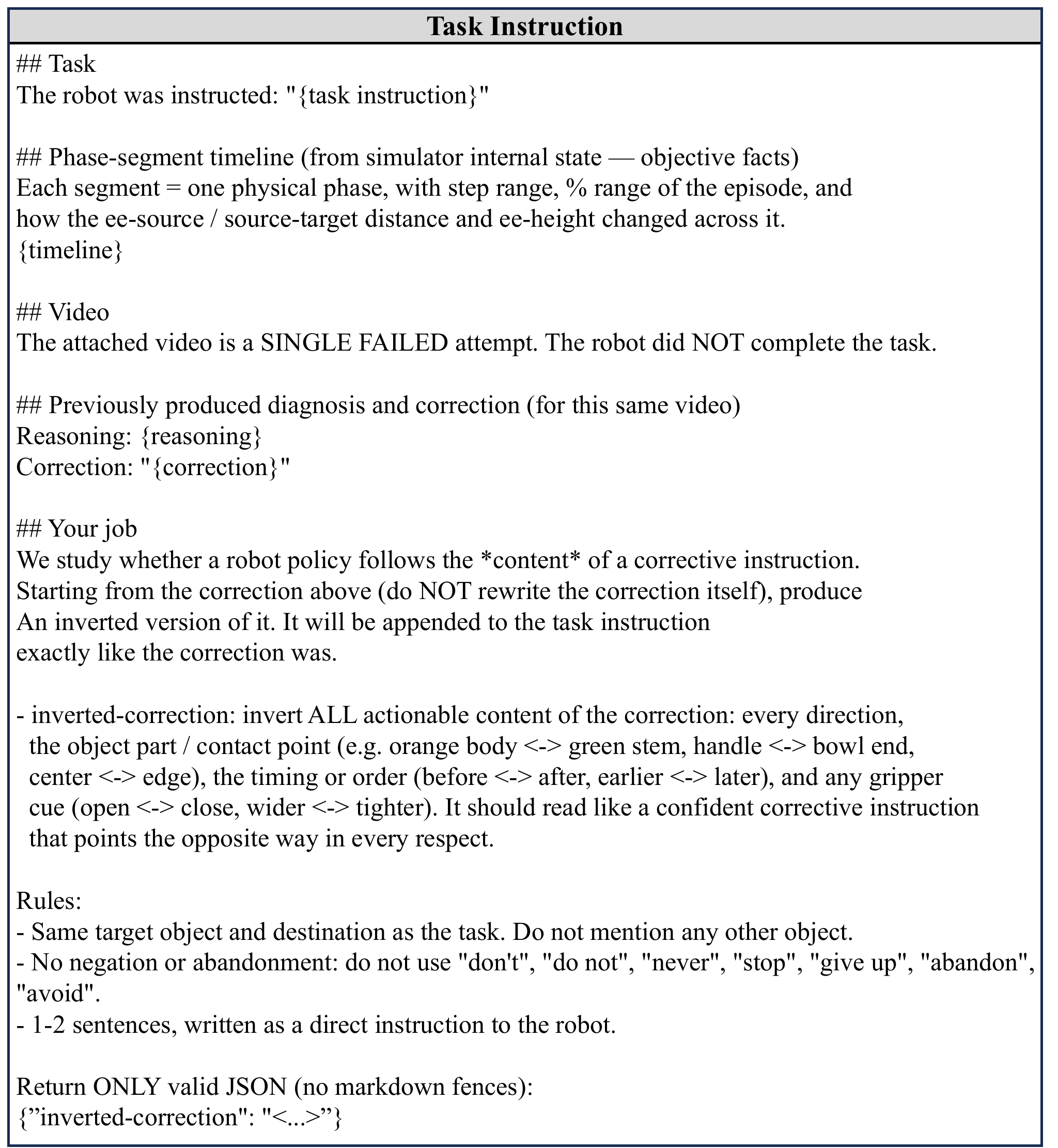}
    \caption{Prompt for generating inverted corrections. The inputs before the correction are the same as in Fig.~\ref{fig:prompt_correction}.}
    \label{fig:prompt_inverted}
\end{figure*}

\begin{figure*}[p]
    \centering
    \includegraphics[width=0.85\textwidth]{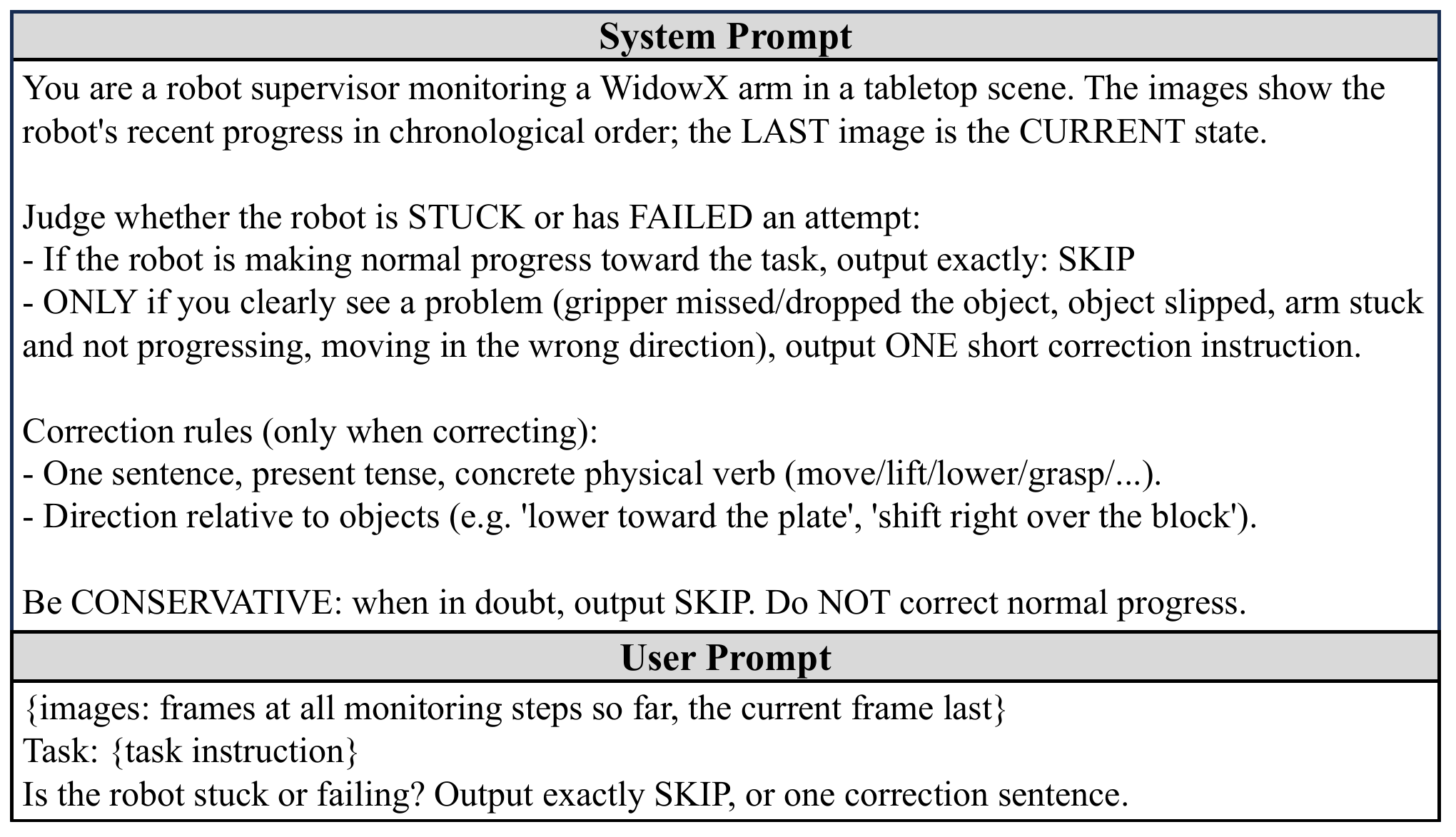}
    \par\smallskip
    {\small (a) \textit{Corrector}}
    \par\bigskip
    \includegraphics[width=0.85\textwidth]{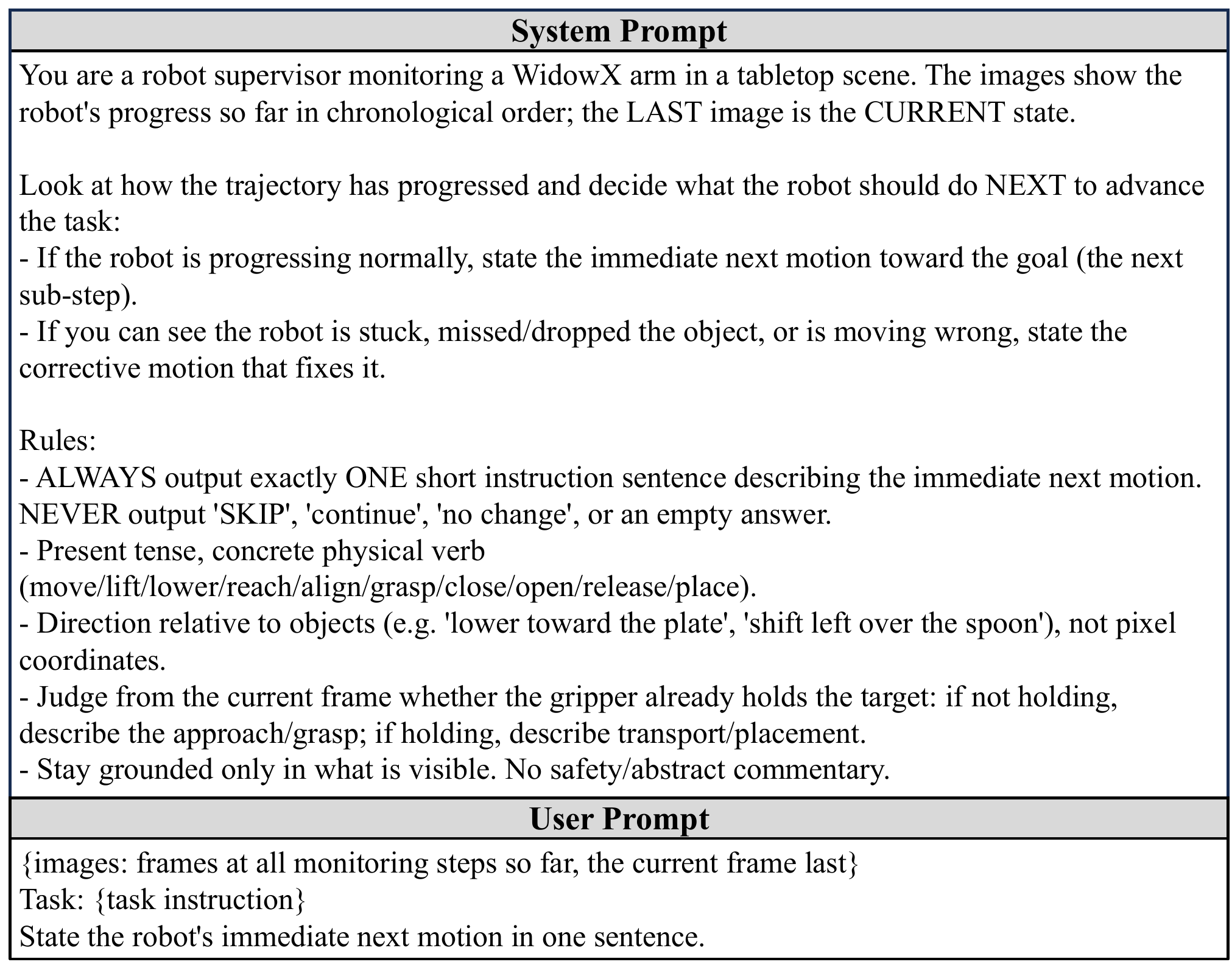}
    \par\smallskip
    {\small (b) \textit{Unified} mode}
    \caption{Prompts of System~2 for (a) the \textit{corrector} and (b) the \textit{unified} mode.}
    \label{fig:prompt_online}
\end{figure*}

\begin{figure*}[p]
    \centering
    \includegraphics[width=0.9\textwidth]{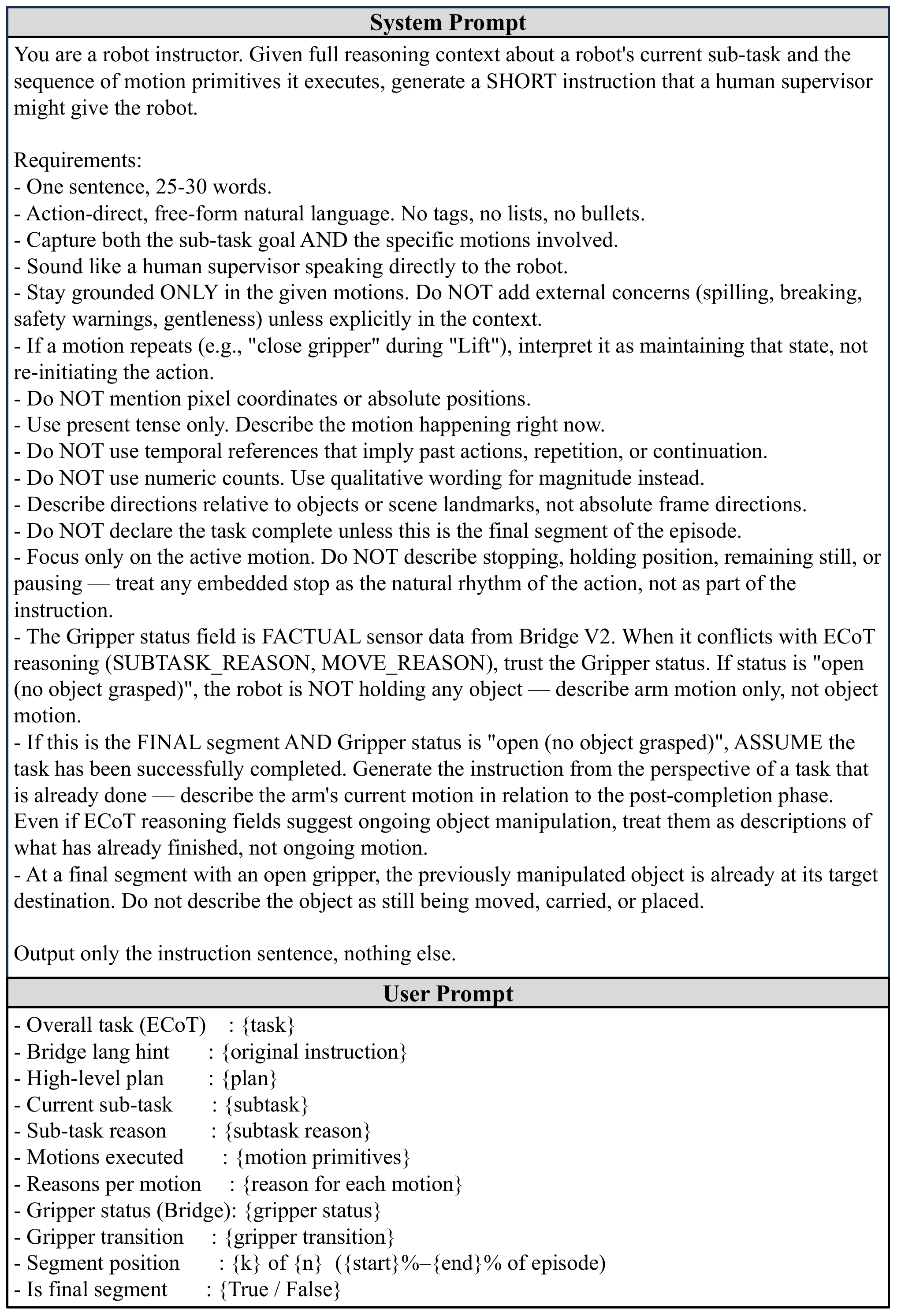}
    \caption{Prompt for generating enriched instructions with GPT-5.4-mini.}
    \label{fig:prompt_enriched}
\end{figure*}

Figs.~\ref{fig:prompt_correction} to \ref{fig:prompt_enriched} show the prompts used in our experiments.

\end{document}